\documentclass[twoside,leqno,twocolumn]{article}
\usepackage{Package/ltexpprt}

\usepackage[font={footnotesize,bf}]{caption}

\usepackage{graphicx}
\graphicspath{ {./Figure/} }
\usepackage{subcaption}

\usepackage{amsmath}
\usepackage{bbm}
\usepackage{mathrsfs}
\DeclareMathOperator*{\argmin}{arg\,min}
\usepackage{amssymb}
\usepackage{upgreek}
\usepackage{mathtools}
\usepackage{lipsum}

\makeatletter
\newcommand{\vast}{\bBigg@{4}}
\newcommand{\Vast}{\bBigg@{5}}
\makeatother

\usepackage{array}
\usepackage{makecell}
\usepackage{tabularx}

\usepackage[algo2e, ruled, linesnumbered]{algorithm2e}

\usepackage{etoolbox}
\makeatletter
\patchcmd{\@makecaption}
{\scshape}
{}
{}
{}
\makeatletter
\patchcmd{\@makecaption}
{\\}
{.\ }
{}
{}
\makeatother

\begin{document}

\title{Detecting and Ranking Causal Anomalies \\ in End-to-End Complex System}

\author{
Ching Chang \\
\and
Wen-Chih Peng
}
\date{National Chiao Tung University, Hsinchu, Taiwan \\ \{blacksnail789521.cs05g, wcpeng\}@nctu.edu.tw}

\maketitle

% Default Copyright Statement
% \fancyfoot[R]{\scriptsize{Copyright \textcopyright\ 2019 by SIAM\\
% Unauthorized reproduction of this article is prohibited}}

% Depending on which copyright you agree to when you sign the copyright form, the copyright 
% can be changed to one of the following after commenting out the default copyright statement
% above.

%\fancyfoot[R]{\scriptsize{Copyright \textcopyright\ 20XX\\
%Copyright for this paper is retained by authors}}

%\fancyfoot[R]{\scriptsize{Copyright \textcopyright\ 20XX\\
%Copyright retained by principal author's organization}}

%\pagenumbering{arabic}
%\setcounter{page}{1}%Leave this line commented out.

\begin{abstract} 
\small
% \baselineskip=9pt

% 隨著科技快速的發展，大型工廠的自動化監控系統也越來越重要。
With the rapid development of technology, the automated monitoring systems of large-scale factories are becoming more and more important.
% 藉由收集大量的機台感測器資料，我們可以有很多方法來找出異常值。
By collecting a large amount of machine sensor data, we can have many ways to find anomalies.
% 我們認為自動化監控系統真正的核心價值就是在找出並追蹤發生問題的原因。
We believe that the real core value of an automated monitoring system is to identify and track the cause of the problem.
% 最著名的尋找causal anomalies的方法為RCA，但是其存在著很多不容忽視的問題。
The most famous method for finding causal anomalies is RCA \cite{cheng2016ranking}, but there are many problems that cannot be ignored.
% 他們利用ARX來建立time-invariant 的correlation network當作機台的profile，然後再拿這個profile用fault propagation的方式來追蹤causal anomalies。
They used the AutoRegressive eXogenous (ARX) model to create a time-invariant correlation network as a machine profile, and then use this profile to track the causal anomalies by means of a method called fault propagation.
% 利用ARX建立的correlation network來描述一個機台的行為有兩大問題: (1)沒有考慮到states的多樣性 (2)沒有把擁有不同time-lag的correlation分開考慮。
There are two major problems in describing the behavior of a machine by using the correlation network established by ARX: (1) It does not take into account the diversity of states (2) It does not separately consider the correlations with different time-lag.
% 基於這些問題，我們提出了一個叫RCAE2E的framework，其徹底解決了上面所述的兩大問題。
Based on these problems, we propose a framework called \textit{Ranking Causal Anomalies in End-to-End System} (RCAE2E), which completely solves the problems mentioned above.
% 實驗的部分，我們分別使用synthetic data以及真實世界的大型光電廠資料來驗證假設的正確性以及存在性。
In the experimental part, we use synthetic data and real-world large-scale photoelectric factory data to verify the correctness and existence of our method hypothesis.

\end{abstract}
\section{Introduction} \label{Introduction}

% 1. 介紹自動監控系統的重要性
% 2. correlation以及建立profile的重要性
% 3. 傳統使用invariant network來找anomaly都不是causal anomaly(需要考慮fault propagation)
% 4. RCA的兩個主要缺點(1. 沒有考慮不同的state 2.沒有考慮到cross time correlation)
% 5. 解決RCA缺點的方法概述
% 6. 本篇論文的主要貢獻

%#####################################################################

%%% 介紹自動監控系統的重要性 %%%
%%% 傳統使用correlation network來找anomaly都不是causal anomaly(需要考慮fault propagation) %%%

% 現今的工廠為了達到自動化的監控系統，其機台都裝設多個sensor。
Nowadays, in order to achieve an automated surveillance system, large-scale factories are equipped with multiple sensors.
% 有了自動化的監控系統以後，我們就可以更方便的地去監控各個機台的執行狀態。
With an automated surveillance system, it is easier to monitor the execution status of each machine.
% 在過往有很多人做了自動檢測複雜機台的異常值。
In the past, many people have done the automatic detection of complex machine anomalies.
% Gertler提出了利用各個sensor的各自行為建議各個限制條件，如果超過了那個條件就會有警示。
Gertler et al. \cite{gertler1997fault} proposed using each sensor's respective behavior to suggest various restrictions, and if it exceeds that condition, there will be a warning.
% 然而，如果各個sensor都單獨考慮的話，會沒有考慮到不同sensor之間的correlations。
However, if each sensor is considered separately, the correlations between different sensors will not be considered.
% 因此，現今的方法都傾向於為系統建立profile來保存correlation的資訊。
Therefore, today's methods \cite{jiang2006discovering} tend to create a profile for the system to preserve the information of the correlations.
% correlation network是個用來表達機台profile最著名的方法之一。
The correlation network is one of the most famous methods used to express machine profiles \cite{jiang2006discovering}.
% 在Correlation network裡，node代表著sensor，而edge代表著兩個sensor之間的correlation。
In the correlation network, a node represents a sensor and an edge stands for a correlation between two sensors.
% 若兩個sensor間有很強的correlation存在的話，代表這兩個sensor的資料具有很大程度的直接關係。
If there is a strong correlation between the two sensors, it means that data from these two sensors have a strong direct relationship.
% 在Ge以及Tao他們各自所提出的方法中，這兩個方法都是利用這些correlations來檢測異常的node。
In Ge et al. \cite{ge2014ranking} and Tao et al. \cite{tao2014metric}'s respective proposed methods, both used these correlations to detect anomalous nodes.
% 他們利用觀察correlations的破裂比例來定義該node是否有異常。
They observed the percentage of broken edges (correlations) to determine whether the node is anomalous.
% 這個方法有明顯的缺點，就是沒有考慮到fault propagation的性質。
The obvious disadvantage of these methods is that it did not take into account the nature of fault propagation.
% 在Wei Cheng所提出的方法，Ranking Causal Anomalies (RCA)，他們充分的考慮了fault propagation的情況。
In the method proposed by Cheng et al. \cite{cheng2016ranking}, called \textit{Ranking Causal Anomalies} (RCA), they fully considered the situation of fault propagation.
% 他們聲稱因為很少有情況是系統的錯誤是被孤立在那邊的，所以基本上不太可能會發生異常的node，其周遭的node都沒有受到影響。
They claimed that because there are few cases in which the system errors are isolated, it's unlikely that the nodes around the anomalous node are not affected.
% 如果有一個node發生異常了，那麼與那個異常node有強烈correlation的node也會接連受到影響。
If there is a node becomes anomalous, then the nodes that have strong correlations with that anomalous node will be affected in succession.
% 因為如此，所有找出來的異常值都會是源頭的異常值。
Because of this, all the anomalies they found will be the source of anomalies.

%---------------------------------------------------------------------------------------------------

%%% RCA的兩個主要缺點(1. 沒有考慮不同的state 2.沒有把擁有不同time-lag的correlation分開考慮) %%%

% 然而，RCA有兩個很大的缺陷。
However, RCA \cite{cheng2016ranking} has two major drawbacks.
% RCA的第一個主要缺陷是他們只利用一個time-invariant model來描述一個機台的行為，但是現實世界中的機台不會只有單一的狀態。
The first major drawback in RCA is that they used a single time-invariant model to describe the behavior of a machine, which means they assumed the behavior of the machine will not change from beginning to end.
% 我們對於這個觀點提出了強烈的懷疑，因為大部分現實世界中的機台，其行為並不能用一個單一個time-invariant model去衡量。
We have raised a strong suspicion about this view because most real-world machines cannot be measured by a single time-invariant model.
% 我們認為我們需要針對機台的各個state分別建一個correlation network，而不是一個機台建立一個correlation network。
Instead of building a correlation network for each machine, we believe that we need to build a correlation network for each state of the machine.
% RCA的第二個主要缺陷是他們並沒有把擁有不同time-lag的correlation分開考慮。
The second major drawback in RCA is that they did not separately consider the correlations with different time-lag.
% 雖然使用ARX來建立的correlation network可以包含不同time-lag的correlation，但是並沒有辦法分開來考慮。
Although the correlation network established using the ARX model can contain the correlations with different time-lag, we have no way to consider these correlations separately.
% 我們認為把擁有不同time-lag的correlation分開考慮可以更精確的來模擬fault propagation。
We believe that separately considering the correlations with different time-lag can simulate fault propagation more accurately.

%---------------------------------------------------------------------------------------------------

\begin{figure}
    \centering
    \includegraphics[width=0.8\columnwidth]{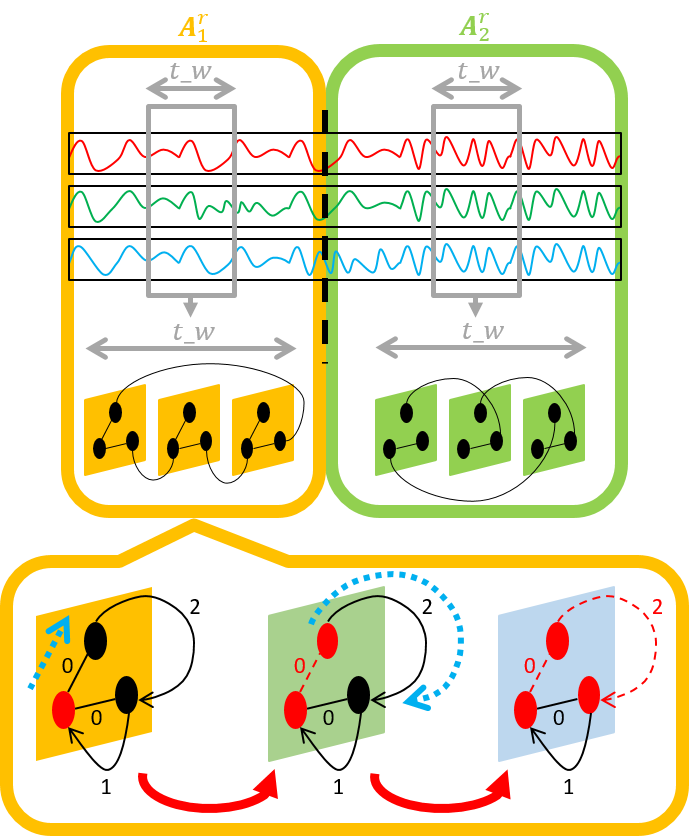}
    \caption{
                % RCAE2E包含了兩個核心方法，第一個是建立profile的TICC_GTC(圖的上半部)，第二個是尋找causal anomalies的RCA_SCC(圖的下半部)。
                RCAE2E includes two core methods. 
                The first is TICC\_GTC for profile creation (upper half of the figure), and the second is RCA\_SCC for ranking causal anomalies (bottom half of the figure).
            }
    \label{RCAE2E}
\end{figure}

%%% 解決RCA缺點的方法概述 %%%

% 為了解決RCA的兩大缺陷，我們提出了一個建立profile的方法叫做TICC_GTC，其引用自David Hallac的方法。
In order to solve the two major drawbacks of RCA, we propose a method to create a profile called Toeplitz Inverse Covariance-based Clustering with Global Temporal Consistency (TICC\_GTC), which is derived from Hallac et al.'s method \cite{hallac2017toeplitz}.
% 他們提出的方法，TICC，可以同時對於多變量時序性資料做segmentation以及clustering。
Their proposed method, Toeplitz Inverse Covariance-based Clustering (TICC), can simultaneously perform segmentation and clustering on multivariate time series data.
% 我們可以利用TICC的segmentation來解決RCA沒有考慮到states的多樣性的問題。
We can use TICC's segmentation to solve the problem that RCA does not consider the diversity of states.
% 我們也可以利用TICC的clustering來解決RCA沒有把擁有不同time-lag的correlation分開考慮的問題。
We can also use TICC's clustering to solve the problem that RCA does not separately consider the correlations with different time-lag.
% 我們所提出的TICC_GTC與TICC唯一不同的地方在於我們額外新增了一個叫global temporal consistency的限制。
The only difference between our proposed TICC\_GTC and TICC is that we have added a new constraint called global temporal consistency.
% 針對於大型工廠內機台的應用，我們認為相鄰run的cluster assignments應該是要對齊的，這是因為工廠生產的行為基本上是非常具有規律的。
For the application of large-scale factory machine, we believe that the cluster assignments of adjacent runs should be aligned because the behavior of factory production is basically very regular.
% 所以說，我們可以利用global temporal consistency去鼓勵相同timestamp不同run的data，其cluster assignment要相同。
So, we can use the global temporal consistency to encourage two data points, which have the same timestamp and adjacent run number, to have the same cluster assignment.
% 在使用TICC_GTC產生完profile後，我們使用了我們提出的方法RCA_SCC來尋找causal anomalies。
After generating the profile by TICC\_GTC, we use our proposed method Ranking Causal Anomalies with Separate Consideration of the Correlations with different time-lag (RCA\_SCC) to find the causal anomalies.
% 顧名思義，RCA_SCC就是把擁有不同time-lag的correlation分開考慮的RCA。
As the name implies, RCA\_SCC is the version of RCA that separately considers the correlations with different time-lag.

%---------------------------------------------------------------------------------------------------

%%% 本篇論文的主要貢獻 %%%

% 總結來說，基於現有方法的局限性，我們提出了一種工廠機器causal anomaly detection的演算法，我們稱之為Ranking Causal Anomalies in End-to-End System (RCAE2E)。
In conclusion, based on the limitations of current methods, we propose a causal anomaly detection algorithm for the factory machine, which we call \textit{Ranking Causal Anomalies in End-to-End System} (RCAE2E).
% Figure 1是RCAE2E的示意圖。
Figure \ref{RCAE2E} is a schematic diagram of RCAE2E.
% 我們的貢獻可以總結為以下。
Our contributions can be summarized as follows.
\begin{enumerate}
    
    % 我們提出了RCAE2E(一個可以在端到端系統裡直接尋找causal anomalies的framework)。我們徹底解決了RCA中沒有考慮多個states以及沒有把擁有不同time-lag的correlation分開考慮的問題。
    \item We propose RCAE2E (a framework that can directly look for causal anomalies in end-to-end system). We completely solve the problem of not considering the diversity of states and not separately considering the correlations with different time-lag in RCA.
    
    % 我們提出了TICC_GTC(一個建立profile的方法)。我們新增了global temporal consistency在TICC上面，使其更符合工廠機台的應用。
    \item We propose TICC\_GTC (a method to build a profile). We add global temporal consistency on TICC to make it more suitable for factory machine applications.
    
    % 我們提出了RCA_SCC(一個尋找causal anomalies的方法)。我們在尋找causal anomalies的時候有把擁有不同time-lag的correlation分開考慮。
    \item We propose RCA\_SCC (a method to find causal anomalies). We separately consider the correlations with different time-lag when looking for causal anomalies.
    
\end{enumerate}
\section{preliminaries and problem definition} \label{preliminaries and problem definition}

% 1. RCA
% 2. TICC
% 3. Problem Definition

\begin{table*}[ht]
\caption{Notations}
\label{Notations}
\centering
\begin{tabular}{ |c|c|c| }
% \begin{tabular}{ |c{1cm}|c{4cm}|c{1cm}| }
\hline
\textbf{Symbol} & \textbf{Definition} & \textbf{Dimension}\\

\hline

$\textbf{x}^{t^r}$ & the data subsequence at time $t-(t\_w-1)$ to time $t$, run $r$ & $\mathbb{R}^{N \times t\_w}$\\

\renewcommand{\arraystretch}{1}

$T \, (R)$ & the length of single run data (the number of run data) & $\mathbb{R}$\\

$N$ & the number of sensors & $\mathbb{R}$\\

\hline

$t\_w \, (r\_w)$ & the window size of multiple data points (multiple run data) & $\mathbb{R}$\\

$K$ & the number of clusters in TICC\_GTC & $\mathbb{R}$\\

$\lambda, \beta, \alpha \, (c, \xi)$ & the parameters in TICC\_GTC (RCA\_SCC) & $\mathbb{R}$\\

\hline

$\textbf{A}_k^r$ & the adjacency matrix of MRF for cluster $k$ from run $r-(r\_w-1)$ to run $r$ & $\mathbb{R}^{(N \cdot t\_w) \times (N \cdot t\_w)}$\\

\hline

$\textbf{b}^t$ & the propagated anomaly score vector at time $t$ & $\mathbb{R}^{N \times 1}$\\

$\textbf{s}^t$ & the causal anomaly score vector at time $t$ & $\mathbb{R}^{N \times 1}$\\

\hline

$\makecell{\textbf{A}^t \\ (\textbf{B}^t)}$ & the adjacency matrix of (broken) MRF from time $t-(t\_w-1)$ to time $t$ & $\mathbb{R}^{(N \cdot t\_w) \times (N \cdot t\_w)}$\\

$\makecell{\textbf{A}^{t,\,(time\_lag)} \\ (\textbf{B}^{t,\,(time\_lag)})}$ & \makecell{ the (broken) correlation network from time $\uptau$ to time $\uptau + time\_lag$ \\ ($t-(t\_w-1) \leq \uptau \leq t-time\_lag$)} & $\mathbb{R}^{N \times N}$\\
\hline
\end{tabular}
\end{table*}

%#####################################################################

% 在這個section，我們先介紹與本論文最相關的兩個方法，分別是Ranking Causal Anomalies (RCA)跟Toeplitz Inverse Covariance-based Clustering (TICC)。
In this section, we first introduce two methods that are most relevant to this paper: Ranking Causal Anomalies (RCA) \cite{cheng2016ranking} and Toeplitz Inverse Covariance-based Clustering (TICC) \cite{hallac2017toeplitz}.
% 最後，我們會描述本篇論文的問題定義。
In the end, we will describe the problem definition for this paper.

%#####################################################################

\subsection{Ranking Causal Anomalies (RCA)} \label{RCA}

% RCA的目的就是找出correlation network中的causal anomaly。
The purpose of RCA is to find causal anomalies in the correlation network.
% 首先他們在training state利用歷史資料找出correlation network。(這裡他們是用ARX model來建立的)
First, they used the historical data to find the correlation network in the training state. (Here they used the ARX model to build it.)
% 然後在testing state的時候，他們檢查新進來的資料是否也會有存在這些在correlation network裡的correlations。
And then in the testing state, they checked if the new incoming data will also have these correlations in the correlation network.
% 如果correlation消失了，就代表透過此correlation連接的兩個sensors很有可能有問題。
If the correlation disappears, it means that the two sensors connecting through this correlation are likely to have some problems.
% 因此，他們利用消失的correlations來回推sensors的損壞程度(稱為Reconstruction Error)，並且利用label propagation technique來往前求出anomaly的源頭(稱為Fault Propagation)。
Therefore, they used vanishing correlations to reconstruct propagated anomaly scores (called \textit{Reconstruction Error}), and used the label propagation technique \cite{zhou2004learning} to determinate the source of the anomalies (called \textit{Fault Propagation}).
% 我們將會介紹這兩個RCA的主要觀念。
We will introduce these two concepts of RCA.

%#####################################################################

\subsubsection{Fault Propagation} \label{RCA Fault Propagation}

% 在RCA裡面有個主要的假設，就是system faults很少會被隔離。
There is a major assumption in RCA that system faults are rarely isolated.
% 他們認為當某個node有出現問題，那個問題會傳給附近的nodes。
They thought that when a node has a problem, that problem will propagate to the nearby nodes.
% 而傳輸的路徑就會是sensor間的correlations。
And the propagation path will be the correlations between the nodes.
% 他們使用了兩個vector來描述這個propagation process。
They used two vectors to describe this propagation process.
% 第一個是ranking score vector of causal anomalies (s)，其代表了每個sensor是否為casual anomaly。
The first one is causal anomaly score vector ($\textbf{s}$), which represents whether each sensor is a casual anomaly.
% 第二個是propagated anomaly score vector (b)，其代表了fault propagation過後每個sensor的anomalous system status。
The second one is propagated anomaly score vector ($\textbf{b}$), which records the anomalous system status of each sensor after fault propagation.
% 由s傳輸到b的整個propagation process程可以由下列最佳化問題所描述
The entire propagation process from $\textbf{s}$ to $\textbf{b}$ can be described by the following optimization problem
\begin{equation} \label{RCA Fault Propagation objective function}
\min_{\textbf{b} \geq \textbf{0}} c \overbrace{ \sum_{i, j=1}^{N} A_{ij} {\bigg\vert \bigg\vert \frac{b_i}{\sqrt{D_{ii}}} - \frac{b_j}{\sqrt{D_{jj}}} \bigg\vert \bigg\vert}^2 }^\text{smoothness constraint} + (1-c) \overbrace{ \sum_{i=1}^{N} \vert \vert b_i - s_i \vert \vert^2 }^\text{fitting constraint}.
\end{equation}
% 這個目標函數裡面有主要兩個部分。
There are two main parts inside this objective function.
% 第一部分為smoothness constraint，代表著他們希望一個node的anomalous system status會跟其附近的node類似。
The first part is \textit{smoothness constraint}, which means that they hoped the node's anomalous system status would be similar to its nearby nodes.
% 第二部分為fitting constraint，代表著他們希望整個system的最終狀態會跟初始狀態類似。
The second part is \textit{fitting constraint}, meaning that they hoped the final state of the entire system would be similar to the initial state.

%#####################################################################

\subsubsection{Reconstruction Error} \label{RCA Reconstruction Error}

% reconstruction error的想法是利用anomalous system status b來重建broken network。
The idea of reconstruction error is to use propagated anomaly score vector $\textbf{b}$ to reconstruct the broken network.
% 我們前面有提到說，他們在testing state的時候會針對每一個新進來的資料建立相對應的broken network。
We mentioned earlier that they created a corresponding broken network for each new incoming data in the testing state.
% broken network就是用來記錄兩兩sensors間的broken correlation。
The broken network is used to record the broken correlations between two sensors.
% broken correlation意味著用其相連的兩個sensors中，至少有一個有異常。
A broken correlation means that at least one of these two sensors connected by this edge is anomalous.
% 所以兩個sensors的anomalous system status b相乘一定會很大。
Therefore, the product of the two sensors' propagated anomaly score must be large.
% 又因為在broken network中，值很大的edge就代表是broken correlation。所以我們可以利用兩個sensors的anomalous system status r相乘來逼近broken network中的值。
And because a large value of an edge in the broken network represents a broken correlation, we can use the product of the two sensors' propagated anomaly score to approximate the values in the broken network.
% reconstruction error的objective function如下
The objective function of reconstruction error is
\begin{equation} \label{RCA Reconstruction Error objective function}
\argmin_{\textbf{b} \geq \textbf{0}} \sum_{i, j=1}^{N} ( b_i \cdot b_j \cdot M_{ij} - \widetilde{B}_{ij} )^2 .
\end{equation}

%#####################################################################

\subsection{Toeplitz Inverse Covariance-based Clustering (TICC)} \label{TICC}

% TICC的目的就是為了可以使多變量的時序性資料切割成多個interpretable的states或者是clusters。
The purpose of TICC is to segment multivariate time series data into multiple interpretable states or clusters.
% 在TICC裡，每個cluster都是一個利用MRF所描述的correlation network。
In TICC, each cluster is a correlation network described by Markov Random Field (MRF).
% MRF可以有很多層，也就是說correlation network可以包含擁有不同time-lag的correlations。
MRF can have many layers, which means that the correlation network can contain the correlations with different time-lag.
% MRF的層數等於window size，也就是一開始所定義的一個time interval，其代表一個timestamp需要考慮的time point數量。
The number of MRF's layers equals to the window size, which is a time interval representing the number of time points we need to consider at one timestamp.
% 他們根據計算sparse Gaussian inverse covariance matrix來學習每個MRF，而每個MRF都可以用一個adjacent matrix A來表示。
They learned each MRF based on the calculation of the sparse Gaussian inverse covariance matrix \cite{friedman2008sparse}, and each MRF can be represented by a adjacency matrix $\textbf{A}$ \cite{banerjee2008model}.
% 當A i,j很大時，就代表sensor i跟sensor j的關係很強。
When $A_{i,j}$ is large, it means that the partial correlation between sensor $i$ and sensor $j$ is very strong.
% 現在，為了讓單一個MRF就可以代表整個cluster，所以他們額外在graphical lasso加上了block Toeplitz這個constraint。
Now, in order to represent the entire cluster with an single MRF, they added the block Toeplitz constraint \cite{gray2006toeplitz} into graphical lasso.
% 因此這樣就可以保證任何在layer i跟layer i+t的edge，也一定會存在在layer i+k跟layer i+k+t裡面。
Therefore, it is guaranteed that any edge between layer $i$ and layer $i+t$ will also exist in layer $i+k$ and layer $i+k+t$.
% 這代表在這個cluster裡面，我們可以用這個MRF來描述所有的data subsequences。
It means that in this cluster, we can use this MRF to describe all the data subsequences.
% TICC的output就是多維度時序性資料的profile，其包含了代表每個cluster的MRF以及每個data subsequence的分配狀態。
The output of TICC is the profile of multivariate time series data, which includes the MRFs ($\textbf{A}=\{\textbf{A}_k : 1 \leq k \leq K\}$) representing each cluster and the cluster assignments ($\textbf{P}=\{\textbf{P}_k : 1 \leq k \leq K\}$) of each data subsequence.
% TICC的整體優化問題為
The overall optimization problem of TICC is 
\begin{equation} \label{TICC objective function}
\begin{split}
\argmin_{\textbf{A} \in \textbf{T},\textbf{P}} \sum_{k=1}^{K} \Bigg[ \overbrace{ \vert \vert \lambda \circ \textbf{A}_k \vert \vert_1 }^\text{sparsity}  + \sum_{\textbf{x}^t \in \textbf{P}_k} \bigg( & - \overbrace{ \ell\ell \big( \textbf{x}^t, \textbf{A}_k \big) }^\text{log likelihood}
\\ 
& + \overbrace{ \beta \mathbbm{1} \big\{ \textbf{x}^{t-1} \notin \textbf{P}_k \big\} }^\text{temporal consistency} \bigg) \Bigg].
\end{split}
\end{equation}
% A屬於T代表著找到的MRF一定要有block Toeplitz structure。
$\textbf{A} \in \textbf{T}$ means the MRF we found must has the block Toeplitz structure.
% 這個目標函數裡面有主要三個部分。
There are three main parts inside this objective function.
% 第一部分為sparsity，代表著他們希望MRF可以很sparse。
The first part is \textit{sparsity}, which means that they hoped the MRFs can be very sparse.
% 根據以往經驗我們知道sparse graphical representations是一個很有效的避免overfitting的方法。
Based on past experience, we know that sparse graphical representations are a very effective way to avoid overfitting \cite{lauritzen1996graphical}.
% 第二部分為log likelihood，代表著data subsequence x^t來自於cluster i的可能性。
The second part is \textit{log likelihood}, which means the likelihood that data subsequence $\textbf{x}^t$ comes from cluster $i$.
% log likelihood的等式為
The equation for log likelihood is
\begin{equation} \label{TICC_log_likelihood}
\begin{split}
\ell\ell(\textbf{x}^t, \textbf{A}_k) = & -\frac{1}{2}(\textbf{x}^t - \mu_k)^T \textbf{A}_k(\textbf{x}^t - \mu_k)
\\
& + \frac{1}{2} \log \, \det \textbf{A}_k - \frac{n}{2} \log (2\pi).
\end{split}
\end{equation}
% 第三部分為temporal consistency，代表著他們希望相鄰兩個data subsequence的cluster assignments應該要一樣。
The third part is \textit{temporal consistency}, which means that they wanted the cluster assignments of two adjacent data subsequences to be the same.
% 如果目前的data subsequence跟上一個data subsequence的cluster assignment不一樣的話，那麼就會有一個penalty。
If the cluster assignment of current data subsequence is not the same as the previous data subsequence, there will be a penalty.

%#####################################################################

\subsection{Problem Definition} \label{Problem Definition}

% RCAE2E的輸入資料為多個run的多變量時序性資料。
The input data of RCAE2E is multivariate time series data with multiple runs.
% 我們用X來表示輸入資料:
We use \textbf{X} to indicate input data: $\textbf{X}=\{x_n^{t^r}:1 \leq n \leq N, 1 \leq t \leq T, 1 \leq r \leq R \}$, where N is the number of sensors, T is the length of single run data, and R is the number of run data.
% 不過在分析輸入資料的時候，我們不是逐個timestamp單獨的看。
However, when analyzing the input data, we do not look at each timestamp individually.
% 取而代之的是，我們是看一個short subsequence，其window size為w。
Instead, we look at a short subsequence, which window size is $t\_w \ll \textbf{T}$.
% 原因是因為我們在建立機台的行為時，我們必須考慮我們分析的資料的上下文。
The reason is that when we establish the behavior of the machine, we must consider the context of the data we analyzed.
% 如此一來我們就可以對於資料有更多的了解。
As a result, we can have more understanding of the data.
% RCAE2E的輸出資料為在有問題的run data裡面，每個data point的每個sensor的causal anomaly score，其值越高代表這個sensor越有可能是causal anomaly。
The output data of RCAE2E is the causal anomaly score for each sensor of each data point in the anomalous run data, and the higher value means the more likely this sensor is a causal anomaly.
% 我們用Y來表示輸出資料:
We use \textbf{Y} to indicate output data: $\textbf{Y}=\{ s_n^t:1 \leq N, 1 \leq t \leq T \}$.
\section{Framework of RCAE2E} \label{Framework of RCAE2E}

\begin{figure}
    \centering
    \includegraphics[width=0.9\columnwidth]{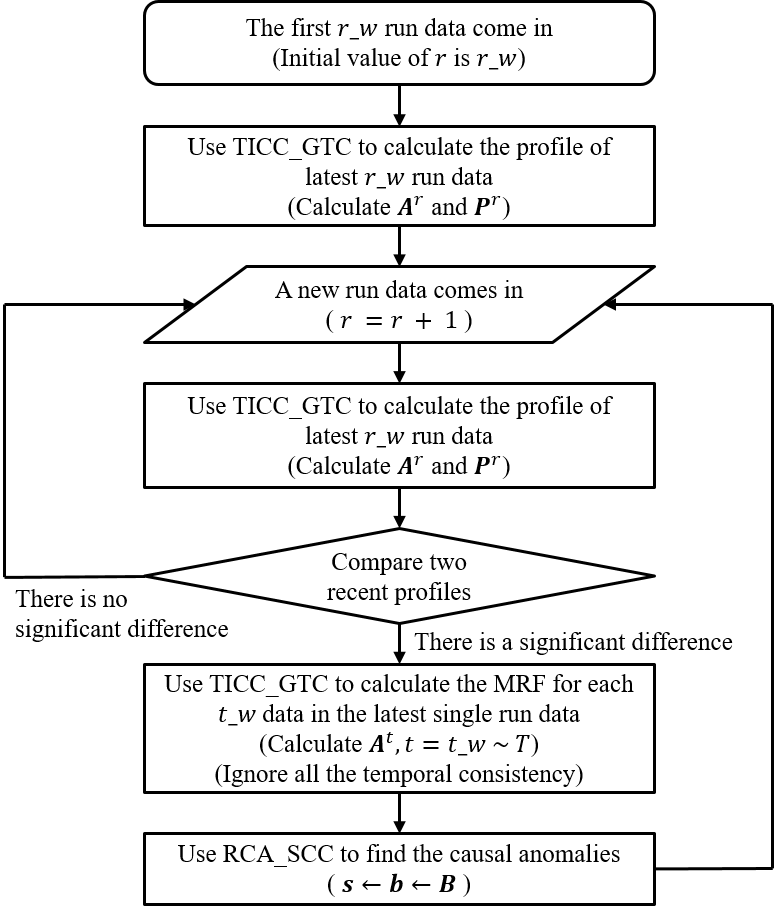}
    \caption{Flowchart of RCAE2E}
    \label{Flowchart}
\end{figure}

% 在這個section，我們詳細介紹了RCAE2E的framework。
In this section, we present the framework of RCAE2E.
% 我們會使用flow chart來描述framework，讓整個過程可以更好理解。
We use a flowchart (Figure \ref{Flowchart}) to describe the framework so that the whole process can be better understood.
% 所有notations都可以在Table X找到。
All notations can be found in Table \ref{Notations}.

%---------------------------------------------------------------------------------------------------

% 在執行RCAE2E的一開始，我們需要有r_w個run data。
At the beginning of the execution of RCAE2E, we need to have $r\_w$ run data.
% r是要用來記錄最新run data的run number的index，而每次利用TICC_GTC計算這r_w個run data的profile可以得到A^r以及P^r。
$r$ is the index used to record the run number of the latest run data, and we can get $\textbf{A}^r=\{\textbf{A}^r_k : 1 \leq k \leq K\}$ and $\textbf{P}^r=\{\textbf{P}^r_k : 1 \leq k \leq K\}$ every time we use TICC\_GTC to calculate the profile of $r\_w$ run data.
% 每當有新的run data進來，我們就會做三件事情：(1)r加1。 (2)把上一次計算的profile標註為old profile。 (3)計算最新的r_w個run data的profile並標註為new profile。
Whenever there is a new run data coming in, we will do three things: (1) $r$ plus 1. (2) Mark the last calculated profile as old profile. (3) Calculate the profile of the latest $r\_w$ run data and mark it as the new profile.
% 當我們有超過一個的profile(也就是說，new profile跟old profile都有值)，我們就可以進入比較profile的環節。
When we have more than one profile (i.e. both new profile and old profile have values.), we can enter the profiles comparison.
% 我們可以利用比較最新的兩個profile來判斷最新的run data是否有異常。
We can use the comparison of two latest profiles to determine if the latest run data is abnormal.
% 這是因為如果最新的run data有明顯的問題，他就會造成最新的profile會與上一個profile有明顯的差異。
This is because if the latest run data has some severe anomalies, it will cause the latest profile to be significantly different from the previous profile.
% 如果我們有找出異常的run data，我們就會進入到下一個階段：尋找causal anomalies。
If we find the anomalous run data, we will move on to the next stage: looking for the causal anomalies.
% 我們會把這個有異常的run data先丟進去TICC_GTC來建立各個data subsequence的MRF，然後再把這些MRF跟ground truth MRF來比較，以求得每個data subsequence的broken MRF。
We feed this anomalous run data into TICC\_GTC to create the MRF of each data subsequence and then compare these MRFs with the ground truth MRFs to get the broken MRF for each data subsequence.
% 值得注意的是，我們在利用TICC_GTC建立有異常的run data裡各個data subsequence的MRF的時候，我們是忽略local temporal consistency的，因為我們需要找到的是可以代表各個data subsequence的MRF。(本來就沒有global temporal consistency，因為現在只有一個run data)
It should be noted that when using TICC\_GTC to create the MRF of each data subsequence in the anomalous run data, we ignore the local temporal consistency because we need to found the MRF that can represent each data subsequence. (There is no global temporal consistency because there is only one run data.)
% 最終，我們把每個data subsequence的ground truth MRF以及broken MRF丟入RCA_SCC，最後就可以得到每個data point(不是data subsequence)的causal anomaly score vector。
Finally, we feed the ground truth MRF and the broken MRF of each data subsequence into the RCA\_SCC, and we can get the causal anomaly score vector for each data point (not the data subsequence).
\section{Detailed Methods of RCAE2E} \label{Detailed Methods of RCAE2E}

% 在這個section，我們介紹包含在RCAE2E裡面的三個詳細方法，其分別是TICC_GTC，Compare Two Profiles，以及RCA_SCC。
In this section, we will introduce three detailed methods included in RCAE2E, which are TICC\_GTC, Compare Two Profiles, and RCA\_SCC.

%######################################################################

\subsection{TICC\_GTC} \label{TICC_GTC}

% 傳統在分析多個run data的時候，我們都是把多個run data串接成一個很長的time series data，然後直接丟到model裡面做分析。
Traditionally, when analyzing multiple run data, we serially connect multiple run data into a long time series data, and then directly feed them into the model for analysis.
% 我們認為這樣的方法是不好的，因為這樣會沒有考慮到run data之間的狀態穩定性質。
We think that this method is not good because this will not take into account the state stability between adjacent run data.
% 一般來說，工廠機台的資料都非常的具有規律性，所以每個run的state分佈應該都會非常類似。
In general, the behavior of factory production is very regular, so the state distribution of adjacent run should be very similar.
% 我們應該要鼓勵兩個data subsequences，其擁有相同的時間點以及相鄰的run number，有相同的cluster assignment。
We need to encourage two data subsequences, which have the same timestamp and adjacent run number, to have the same cluster assignment.
% 我們把這個新的限制稱為global temporal consistency，並且把原本在TICC裡面的temporal consistency改稱為local temporal consistency。
We call this new constraint \textit{global temporal consistency} (GTC), and renamed the original temporal consistency in TICC to \textit{local temporal consistency}.

%---------------------------------------------------------------------------------------------------

% TICC_GTC就是TICC加上GTC，所以原本TICC的objective function會因為加上了GTC而變成以下式子。
TICC\_GTC is TICC plus GTC, so the original objective function of TICC (Problem (\ref{TICC objective function})) will become to
\begin{equation} \label{TICC_GTC objective function}
\begin{split}
& \argmin_{\textbf{A}^r \in \textbf{T},\textbf{P}^r} \sum_{k=1}^{K} \Bigg[ \overbrace{ \vert \vert \lambda \circ \textbf{A}_k^r \vert \vert_1 }^\text{sparsity}  + \sum_{\textbf{x}^{t^r} \in \textbf{P}_k^r} \bigg( - \overbrace{ \ell\ell \big( \textbf{x}^{t^r}, \textbf{A}_k^r \big) }^\text{log likelihood}
\\ 
& + \overbrace{ \beta \mathbbm{1} \big\{ \textbf{x}^{[t-1]^r} \notin \textbf{P}_k^r \big\} }^\text{local temporal consistency} + \overbrace{ \alpha \mathbbm{1} \big\{ \textbf{x}^{t^{r-1}} \notin \textbf{P}_k^r \big\} }^\text{global temporal consistency} \bigg) \Bigg].
\end{split}
\end{equation}
% 我們可以利用類似EM演算法的方法來解Problem X，其在分配data subsequences到clusters以及update MRFs之間交替運行。
We can solve Problem (\ref{TICC_GTC objective function}) by using a variation of Expectation Maximization (EM) algorithm, which alternates between assigning data subsequences to clusters and updating the MRFs.
% 因為跟temporal consistency有關的只有在assign data subsequence到cluster的時候(我們會固定住A)，所以我們可以把Problem X變成
Because the only thing related to temporal consistency is assigning data subsequences to clusters (we will fix $\textbf{A}^r$), we can change Problem (\ref{TICC_GTC objective function}) to
\begin{equation} \label{TICC_GTC objective function for cluster assignments}
\begin{split}
\argmin_{\textbf{P}^r} \sum_{k=1}^{K} \sum_{\textbf{x}^{t^r} \in \textbf{P}_k^r} \bigg( -  \ell\ell \big( \textbf{x}^{t^r}, \textbf{A}_k^r \big) & + \beta \mathbbm{1} \big\{ \textbf{x}^{[t-1]^r} \notin \textbf{P}_k^r \big\}
\\
& + \alpha \mathbbm{1} \big\{ \textbf{x}^{t^{r-1}} \notin \textbf{P}_k^r \big\} \bigg) .
\end{split}
\end{equation}

%---------------------------------------------------------------------------------------------------

\begin{figure} 
    \centering
    
    \begin{subfigure}{0.47\textwidth}
        \includegraphics[width=\textwidth]{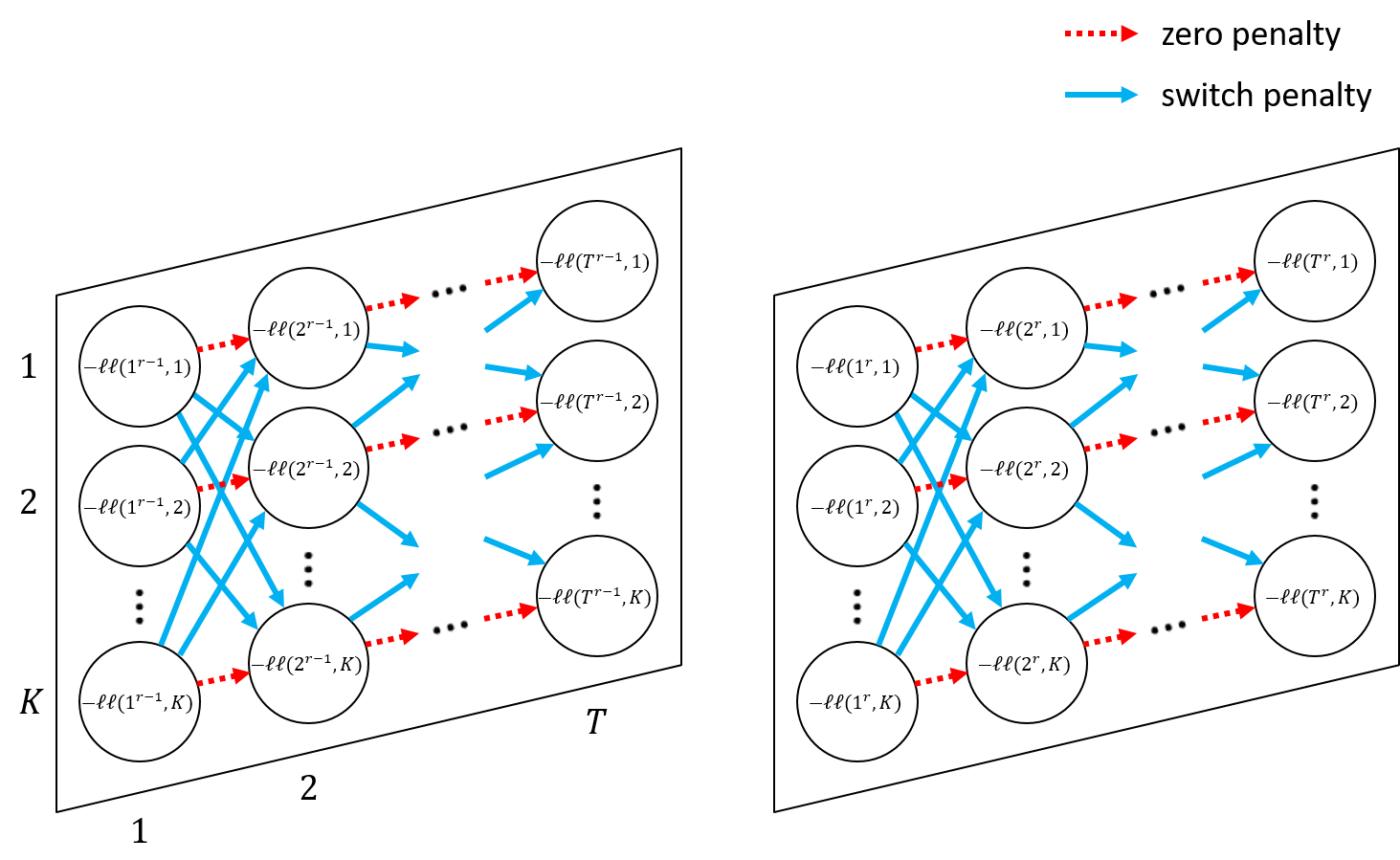}
        \caption{Local Temporal Consistency (LTC)}
        \label{Local Temporal Consistency}
    \end{subfigure}
    
    \begin{subfigure}{0.47\textwidth}
        \includegraphics[width=\textwidth]{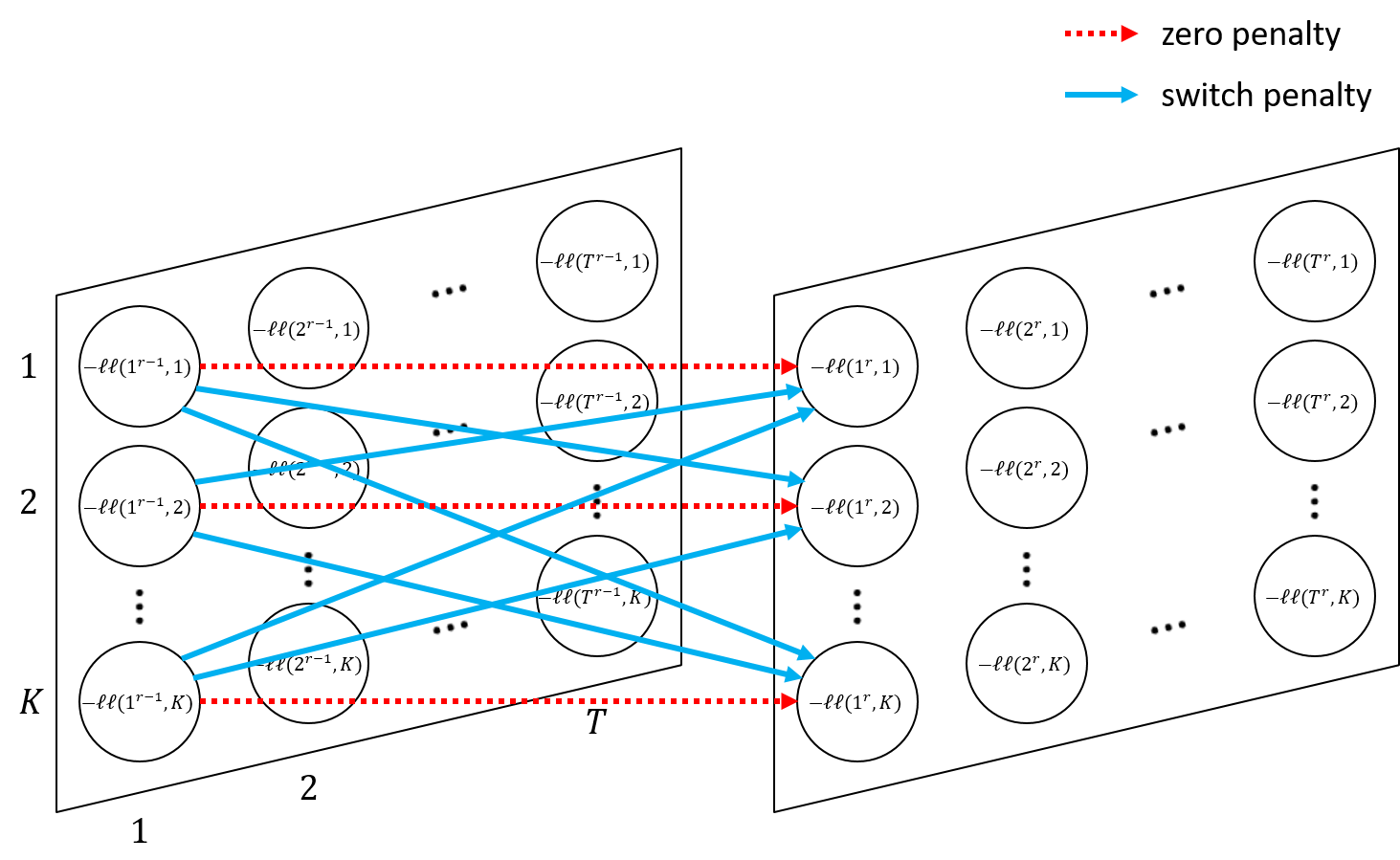}
        \caption{Global Temporal Consistency (GTC)}
        \label{Global Temporal Consistency}
    \end{subfigure}
    
    \caption{Temporal Consistency}
    \label{Temporal Consistency}
\end{figure}

% 在Figure X裡面，node代表data subsequence被assign到相對應的cluster的negative log likelihood，而edge代表了改變cluster assignment的cost。
In Figure \ref{Temporal Consistency}, the node represents the negative log likelihood of the data subsequence being assigned to the corresponding cluster, and the edge represents the cost when switching the cluster assignment.
% 我們必須在每個run裡面選出一條時間從1到T的最短路徑，所以總共會有r_w個路徑。
We need to choose a minimum-cost path from time $1$ to time $T$ in each run, so there are $r\_w$ paths in total.
% 我們利用動態規劃(Algorithm X)來求解Problem X。
We use dynamic programming (Algorithm \ref{Assign Data Subsequences to Clusters}) to solve Problem (\ref{TICC_GTC objective function for cluster assignments}).
% 在Algorithm X裡面，在每個時間戳裡面，我們利用笛卡爾積來列舉每個時間點裡面所有run的cluster assignment(使用AllComb來儲存)，然後對每種combination of cluster assignment找出目前為止的最佳解。
In Algorithm \ref{Assign Data Subsequences to Clusters}, for each timestamp, we use the Cartesian product to enumerate all combinations of each run selecting one cluster (use AllComb to store), and then find the best solution so far for each combination.
% 我們利用解Problem X來算出目前為止的最佳解。這個問題可以表述如下:
We can find the solution by solving Problem (\ref{CurrCost}), which can be formulated as follows: 
\begin{equation} \label{CurrCost}
\begin{split}
& \text{CurrCost[CurrIdx]} =
\\
& \min (\text{PrevCost[PrevIdx]} -\ell\ell(t^r, k) + \text{switch penalty}),
\end{split}
\end{equation}
% switch penalty包括了LTC以及GTC。
where the switch penalty includes LTC and GTC.
% Algorithm X的時間複雜度為XXX。
The time complexity of Algorithm \ref{Assign Data Subsequences to Clusters} is $O(K^{r\_w^2} \cdot r\_w \cdot T)$.
% Algorithm X就是Problem X的pseudo code，2~6行代表LTC，而7~11行代表GTC。
Algorithm \ref{CalCost} is the pseudo code of Problem (\ref{CurrCost}), with lines 2 $\sim$ 6 representing LTC and lines 7 $\sim$ 11 representing GTC.

\SetArgSty{textnormal}
\let\oldnl\nl% Store \nl in \oldnl
\newcommand{\nonl}{\renewcommand{\nl}{\let\nl\oldnl}}% Remove line number for one line
\begin{algorithm2e}
\SetAlgoLined
% \KwResult{Write here the result}
% \textbf{initialize} AllComb = list of all combinations of each run \\ $\qquad \qquad \qquad \qquad \; \; \; \,$selecting one cluster. (length of \\ $\qquad \qquad \qquad \qquad \; \;$ AllComb = $K^{r\_w}$). \\
% $\qquad \qquad$PrevCost = list of length(AllComb) zeros. \\
% $\qquad \qquad$CurrCost = list of length(AllComb) infinite. \\
% $\qquad \qquad$PrevPath = list of length(AllComb) empty lists. \\
% $\qquad \qquad$CurrPath = list of length(AllComb) empty lists. \\
\textbf{given} $\beta > 0, \alpha > 0, -\ell\ell(t^r, k)$. \\
AllComb = list of all combinations of each run selecting one cluster. \\
\nonl \hspace*{1mm} (length of AllComb = $K^{r\_w}$). \\
PrevCost = list of length(AllComb) zeros. \\
CurrCost = list of length(AllComb) infinite. \\
PrevPath = list of length(AllComb) empty lists. \\
CurrPath = list of length(AllComb) empty lists. \\

\For{$t=1, ..., T$}
{
    CurrCost = list of length(AllComb) infinite. \\
    \For{CurrIdx $=1, ..., $ length(AllComb)}
    {
        \For{PrevIdx $=1, ..., $ length(AllComb)}
        {
            CurrCostPlusPrevCost = CalCost(). \\
            \If{CurrCostPlusPrevCost $<$ CurrCost[CurrIdx]}
            {
                CurrCost[CurrIdx] = CurrCostPlusPrevCost. \\
                CurrPath[CurrIdx] = PrevPath[PrevIdx] \\
                \nonl \hspace*{1mm} .append(AllComb[CurrIdx]). \\
            }
        }
    }
    PrevCost = CurrCost. \\
    PrevPath = CurrPath. \\
}
FinalMinIdx = index of minimum value of CurrCost. \\
FinalCost = CurrCost[FinalMinIdx]. \\
FinalPath = CurrPath[FinalMinIdx]. \\
\textbf{return} FinalCost, FinalPath. \\

\caption{Assign Data Subsequences to Clusters}
\label{Assign Data Subsequences to Clusters}
\end{algorithm2e}

% \quad	space equal to the current font size (= 18 mu)
% \,	3/18 of \quad (= 3 mu)
% \:	4/18 of \quad (= 4 mu)
% \;	5/18 of \quad (= 5 mu)
% \!	-3/18 of \quad (= -3 mu)
% \ (space after backslash!)	equivalent of space in normal text
% \qquad	twice of \quad (= 36 mu)
\SetArgSty{textnormal}
\begin{algorithm2e}
\SetAlgoLined
CurrCostPlusPrevCost = PrevCost[PrevIdx] $-\ell\ell(t^r, k)$ corresponding to AllComb[CurrIdx].

\For{$r=1, ..., r\_w$}
{
    \If{$t \neq 0$ \textbf{and} AllComb[PrevIdx][r] $\neq$ AllComb[CurrIdx][r]}
    {
        CurrCostPlusPrevCost += $\beta$. \\
    }
}

\For{$r=1, ..., r\_w$}
{
    \If{$r \neq 0$ \textbf{and} AllComb[PrevIdx][r] $\neq$ AllComb[CurrIdx][r]}
    {
        CurrCostPlusPrevCost += $\alpha$. \\
    }
}

\textbf{return} CurrCostPlusPrevCost. \\

\caption{CalCost}
\label{CalCost}
\end{algorithm2e}

%######################################################################

\subsection{Compare Two Profiles} \label{Compare Two Profiles}

% 當在比較兩個profiles的時候，我們需要做兩件事情。
When comparing two profiles, we need to do two things.
% 第一，我們需要先把這兩個profile的長度都縮減到長度為T，因為目前的profile長度為T*r_w。
First, we need to reduce the length of both profiles to length $T$ because the current profile length is $T \times r\_w$.
% 原因是因為在每個時間點上，我們希望可以找出一個MRF，其可以代表所有run的data subsequence。
The reason is that at each timestamp, we want to find an MRF that can represent the data subsequences of all run data.
% 縮減長度的方法就是把擁有相同時間的MRF都平均起來(每個時間點都會有r_w個MRF)。
So, the method of reducing the length is to average the MRFs with the same timestamp. (There are $r\_w$ MRFs at each timestamp.)
% 第二，我們需要計算這兩個averaged profiles的difference score。
Second, we need to calculate the difference score for these two averaged profiles.
% 我們先計算每個時間點的兩個相對應的MRF的cosine distance(1 - cosine similarity)，然後再把所有timestamp的cosine distance做平均。
We first calculate the cosine distance ($1 - \text{cosine similarity}$) of the two corresponding MRFs at each timestamp and then average the cosine distances of all timestamps.

%######################################################################

\subsection{RCA\_SCC} \label{RCA_SCC}

% 當我們在比較兩個profile發現有明顯差異時，我們就會進入到尋找causal anomalies的階段。
When we compare two profiles and find that there are significant differences, we will enter the stage of finding causal anomalies.
% 我們會把這個有異常的run data先丟進去TICC_GTC來建立各個data subsequence的MRF，然後再把這些MRF跟ground truth MRF來比較，以求得每個data subsequence的broken MRF。
We feed this anomalous run data into TICC\_GTC to create the MRF of each data subsequence and then compare these MRFs with the ground truth MRFs to get the broken MRF for each data subsequence.
% 最後我們才會把每個data subsequence的ground truth MRF以及broken MRF丟進RCA_SCC。
Finally, we feed the ground truth MRF and the broken MRF of each data subsequence into RCA\_SCC.
% RCA_SCC跟RCA的差別就是輸入到RCA_SCC的MRF可以包含擁有不同time-lag的correlations。
The difference between RCA\_SCC and RCA is that the MRFs we feed into RCA\_SCC can contain the correlations with different time-lag.
% 我們會介紹RCA_SCC的兩個部分: 從RCA轉變成RCA_SCC以及得到每個data subsequence的s
Next We will introduce three parts of RCA\_SCC: \textit{Convert from RCA to RCA\_SCC}, \textit{Get the \textbf{s} of Each Data Subsequence} and \textit{Get the \textbf{s} of Each Data Point}.

%######################################################################

\subsubsection{Convert from RCA to RCA\_SCC} \label{RCA_SCC Convert from RCA to RCA_SCC}

% 為了公式簡潔，我們先把RCA裡面的兩個objective function(Fault Propagation以及Reconstruction Error)都轉成向量的形式。
In order to simplify the formula, we first convert the two objective functions (\textit{Fault Propagation} and \textit{Reconstruction Error}) in the RCA into vector form.
% Problem X可以轉成以下式子
Problem (\ref{RCA Fault Propagation objective function}) can be converted into the following formula:
\begin{equation} \label{RCA Fault Propagation objective function vector form}
\min_{\textbf{b} \geq \textbf{0}} c \textbf{b}^T ( \textbf{I}_{N \cdot t\_w}  - \widetilde{\textbf{A}} ) \textbf{b} + (1-c) \vert \vert \textbf{b} - \textbf{s} \vert \vert_F^2 ,
\end{equation}
% 而Problem X可以轉成以下式子
and Problem (\ref{RCA Reconstruction Error objective function}) can be converted into the following formula:
\begin{equation} \label{RCA Reconstruction Error objective function vector form}
\argmin_{\textbf{b} \geq \textbf{0}} \vert \vert ( \textbf{b} \textbf{b}^T ) \circ \textbf{M} - \widetilde{\textbf{B}} \vert \vert_F^2 .
\end{equation}

%---------------------------------------------------------------------------------------------------

\begin{figure}
    \centering
    \includegraphics[width=\columnwidth]{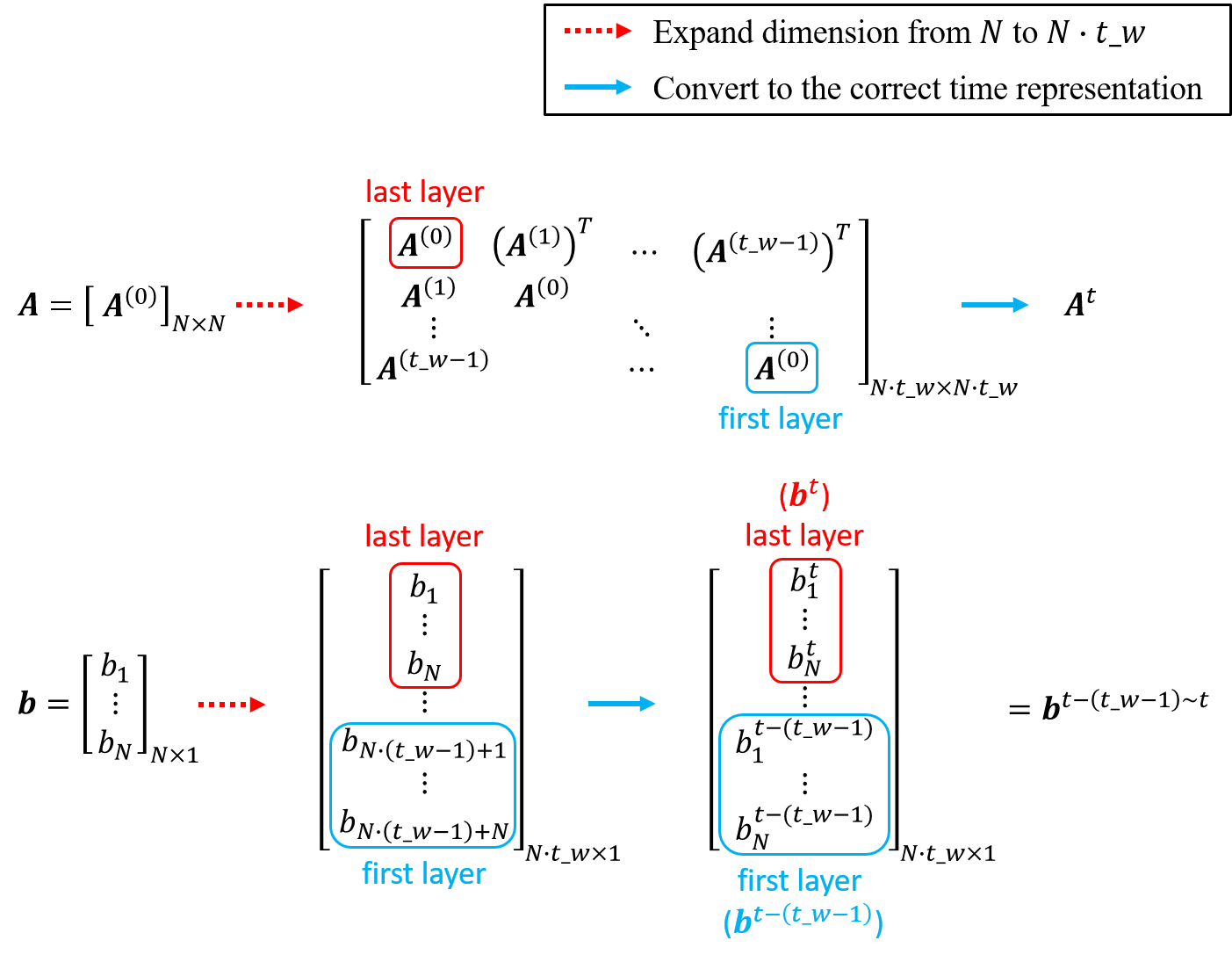}
    \caption{Convert from RCA to RCA\_SCC}
    \label{Convert from RCA to RCA_SCC}
\end{figure}

% 接下來，我們需要把Problem X以及Problem X變成RCA_SCC的格式。
Next, we need to turn Problem (\ref{RCA Fault Propagation objective function vector form}) and Problem (\ref{RCA Reconstruction Error objective function vector form}) into the format of RCA\_SCC.
% RCA轉成RCA_SCC需要兩個步驟：擴展維度以及轉成正確的時間表示。
The conversion of RCA to RCA\_SCC requires two steps: expand the dimension and convert to the correct time representation.
% RCA轉成RCA_SCC的規則可以參考Figure X。
The rules of converting RCA to RCA\_SCC can be referred to Figure \ref{Convert from RCA to RCA_SCC}.
% 因為RCA_SCC所考慮的MRF可以包含擁有不同time-lag的correlations，所以我們所使用的A的維度會比RCA裡面所使用的A大t_w倍 (N X N變成(N*t_w) X (N*t_w))。
Because the MRF considered by RCA\_SCC can contain correlations with different time-lag, the dimension of \textbf{A} we use in RCA\_SCC is $t\_w$ times larger than the \textbf{A} we used in RCA ($\mathbb{R}^{N \times N}$ becomes $\mathbb{R}^{(N \cdot t\_w) \times (N \cdot t\_w)}$).
% 也因為這樣，我們必須把相對應的variable，像是B、b、s，都擴展t_w倍。
Also, we must expand the dimension corresponding variables, such as B, b, s, by $t\_w$ times.
% 然而，光是只有擴展維度是不夠的，因為現在我們只是把t_w個連續的data point視為擁有相同的timestamp。
However, it is not enough to simply expand the dimension, because now we just think of $t\_w$ consecutive data points as having the same timestamp.
% 所以擴展維度後，我們需要把這N*t_w的sensor轉成t_w個連續時間的N個sensor。
So after expanding the dimension, we need to convert these $N \cdot t\_w$ sensors into $N$ sensors of $t\_w$ consecutive timestamps.
% 基於以上的規則，我們可以藉由把Problem X轉變成以下式子來得到RCA_SCC的Fault Propagation的objective function:
Based on the above rules, we can get the objective function of RCA\_SCC's \textit{Fault Propagation} by converting Problem (\ref{RCA Fault Propagation objective function vector form}) into the following formula:
\begin{equation} \label{RCA_SCC Fault Propagation objective function vector form}
\begin{split}
\min_{\textbf{b}^{t-(t\_w-1) \sim t} \geq \textbf{0}}  & c ( \textbf{b}^{t-(t\_w-1) \sim t} )^T ( \textbf{I}_{N \cdot t\_w}  - \widetilde{\textbf{A}^t} ) \textbf{b}^{t-(t\_w-1) \sim t}
\\
& + (1-c) \vert \vert \textbf{b}^{t-(t\_w-1) \sim t} - \textbf{s}^{t-(t\_w-1) \sim t} \vert \vert_F^2 .
\end{split}
\end{equation}
% 也因為這樣，我們可以把Problem X轉變成以下式子:
Also, we can get the objective function of RCA\_SCC's \textit{Reconstruction Error} by converting Problem (\ref{RCA Reconstruction Error objective function vector form}) into the following formula:
\begin{equation} \label{RCA_SCC Reconstruction Error objective function vector form}
\argmin_{\textbf{b}^{t-(t\_w-1) \sim t} \geq \textbf{0}} \vert \vert [ \textbf{b}^{t-(t\_w-1) \sim t} ( \textbf{b}^{t-(t\_w-1) \sim t} )^T ] \circ \textbf{M}^t - \widetilde{\textbf{B}^t} \vert \vert_F^2 .
\end{equation}

%######################################################################

\subsubsection{Get the \textbf{s} of Each Data Subsequence} \label{RCA_SCC Get the s of Each Data Subsequence}

% 我們現在要解兩個RCA_SCC的objective function(分別是Problem X以及Problem X)。
We are now going to solve the two objective functions of RCA\_SCC (Problem (\ref{RCA_SCC Fault Propagation objective function vector form}) and Problem (\ref{RCA_SCC Reconstruction Error objective function vector form}), respectively).
% 解RCA_SCC的Fault Propagation的objective function(Problem X)非常簡單，我們可以直接對b做偏微分來得到closed-form solution。
Solving the objective function of RCA\_SCC's \textit{Fault Propagation}  (Problem (\ref{RCA_SCC Fault Propagation objective function vector form})) is very simple, we can directly differentiate with respect to $\textbf{b}^{t-(t\_w-1) \sim t}$ to get the closed-form solution.
% 所以Problem X的closed-form solution就會是
So the closed-form solution of Problem (\ref{RCA_SCC Fault Propagation objective function vector form}) is \cite{zhou2004learning}
\begin{equation} \label{RCA_SCC Fault Propagation closed-form solution}
\textbf{b}^{t-(t\_w-1) \sim t} =  \textbf{E}^t \textbf{s}^{t-(t\_w-1) \sim t} ,
\end{equation}
where 
\begin{equation} \label{RCA_SCC Fault Propagation closed-form solution E}
\textbf{E}^t = (1-c) ( \textbf{I}_{N \cdot t\_w} - c \widetilde{\textbf{A}}^t )^{-1} .
\end{equation}
% 在得到了Equation X以後，我們可以把RCA_SCC的Reconstruction Error的objective function(Problem X)裡面的b都換成s。
After getting Equation (\ref{RCA_SCC Fault Propagation closed-form solution}), we can replace $\textbf{b}^{t-(t\_w-1) \sim t}$ in the objective function (Problem (\ref{RCA_SCC Reconstruction Error objective function vector form})) of RCA\_SCC's \textit{Reconstruction Error} with $\textbf{s}^{t-(t\_w-1) \sim t}$.
% 另外，因為在系統裡面root cause anomalies的數量通常不會很大量，所以我們對s加上了L1 norm來控制找出來的root cause的數量。
In addition, because the number of root cause anomalies are usually not very large in the system, we added L1 norm to $\textbf{s}^{t-(t\_w-1) \sim t}$ to control the number of root cause anomalies we found.
% 所以，Problem X可以轉成以下式子
So, Problem (\ref{RCA_SCC Reconstruction Error objective function vector form}) can be converted into the following formula:
\begin{equation} \label{RCA_SCC Reconstruction Error objective function only s}
\begin{split}
\argmin_{\textbf{s}^{t-(t\_w-1) \sim t} \geq \textbf{0}} & \vert \vert [ \textbf{E}^t \textbf{s}^{t-(t\_w-1) \sim t} ( \textbf{s}^{t-(t\_w-1) \sim t} )^T ( \textbf{E}^t )^T ] \circ \textbf{M}^t \\
& - \widetilde{\textbf{B}^t} \vert \vert_F^2 + \xi \vert \vert \textbf{s}^{t-(t\_w-1) \sim t} \vert \vert_1 .
\end{split}
\end{equation}
% 我們是利用一個iterative multiplicative updating algorithm來解Problem X的。
We use an iterative multiplicative updating algorithm to solve Problem (\ref{RCA_SCC Reconstruction Error objective function only s}) \cite{cheng2016ranking}.
% 更新的方式為Equation X。
The updating rule is Equation (\ref{RCA_SCC Reconstruction Error updating rule}).
% 我們可以保證利用Equation X去更新的話，每次更新都會使得objective function的值下降。
We can guarantee that if we use Equation (\ref{RCA_SCC Reconstruction Error updating rule}) to update the Problem (\ref{RCA_SCC Reconstruction Error objective function only s}), each update will monotonically decrease the value of the objective function.
% Equation X的時間複雜度為X，Iter代表更新到收斂的次數。
The time complexity of Equation (\ref{RCA_SCC Reconstruction Error updating rule}) is $O(Iter \cdot (N \cdot t\_w)^3)$, where $Iter$ represents the number of updates to convergence.
% 因為版面限制我們這邊忽略了Equation X的推導以及證明其收斂性。
Due to space limitation, we omit the derivation of Equation (\ref{RCA_SCC Reconstruction Error updating rule}) and the proof of its convergence.
% 相關推導以及證明可以參考X。
The relevant derivation and proof can refer to \cite{cheng2016ranking}.

\begin{figure*} [ht]
\begin{equation} \label{RCA_SCC Reconstruction Error updating rule} \tag{4.15}
\textbf{s}^{t-(t\_w-1) \sim t} = \hat{\textbf{s}}^{t-(t\_w-1) \sim t} \circ \Bigg( \frac{4 \{ [ ( \textbf{E}^t )^T \widetilde{\textbf{B}^t} ] \circ \textbf{M}^t \} \textbf{E}^t \hat{\textbf{s}}^{t-(t\_w-1) \sim t} }{4 \{ [ ( \textbf{E}^t )^T \textbf{E}^t \hat{\textbf{s}}^{t-(t\_w-1) \sim t} ( \hat{\textbf{s}}^{t-(t\_w-1) \sim t} )^T ( \textbf{E}^t )^T ] \circ \textbf{M}^t \} \textbf{E}^t \hat{\textbf{s}}^{t-(t\_w-1) \sim t} + \xi \textbf{1}_{N \cdot t\_w}  } \Bigg) ^\frac{1}{4}
\end{equation}
\end{figure*}

%######################################################################

\subsubsection{Get the \textbf{s} of Each Data Point} \label{RCA_SCC Get the s of Each Data Point}

% 這邊要注意的是，Problem X算出來的s是對應到一個data subsequence而不是data point。
It should be noted that the $\textbf{s}^{t-(t\_w-1) \sim t}$ calculated from Problem (\ref{RCA_SCC Reconstruction Error objective function only s}) corresponds to each data subsequence, not each data point.
% 也就是說，每次sliding window，我們算出來的s是包含了t_w個data points的。
In other words, for each sliding window, the $\textbf{s}^{t-(t\_w-1) \sim t}$ we calculated contains $t\_w$ data points.
% 我們最終要求的是每個data point的s，而不是每個data subsequence的s。
What we ultimately ask is the $\textbf{s}^{t}$ of each data point, not the $\textbf{s}^{t-(t\_w-1) \sim t}$ of each data subsequence.
% 所以我們要把算出來的s拆散成t_w個s。
So we have to split the calculated $\textbf{s}^{t-(t\_w-1) \sim t}$ into $t\_w$ $\textbf{s}^{t}$.
% 也因為這樣，所以每個data point會對應到多個s (因為sliding window在移動時是每次移動一個timestamp)。
Because of this, each data point will correspond to multiple $\textbf{s}^{t}$. (Because the distance that each time the sliding window moves is one timestamp.)
% 因此，最終每個data point的s就會是所有對應到那個data point的s的平均。
Therefore, the final $\textbf{s}^{t}$ of each data point will be the average of all $\textbf{s}^{t}$ corresponding to that data point.
\section{Experiments} \label{Experiments}

% 實驗的部分，我們分別使用synthetic data以及真實世界的大型光電廠資料來驗證假設的正確性以及存在性。
In the experimental part, we use synthetic data and real-world large-scale photoelectric factory data to verify the correctness and existence of our method hypothesis.
% 我們選定了一些state-of-the-art的方法，包括了X，X以及X。
We have chosen some state-of-the-art methods, including mRank and gRank \cite{ge2014ranking}, LBP \cite{tao2014metric}, and RCA \cite{cheng2016ranking}.
% 我們選擇RCAE2E以及其他方法的參數的方法是利用BIC以及cross-validation。
The methods we choose RCAE2E and other method parameters are BIC \cite{hastie2009unsupervised} and cross-validation.
% 我們選用的evaluation metrics為precision, recall以及nDCG。
The evaluation metrics we used are precision, recall, and nDCG \cite{jarvelin2002cumulated}.
% precision、recall是利用top-k個我們找到的結果計算出來的，而k通常是真正causal anomalies的數量的兩到三倍左右。
The precision, recall is calculated using the top-$k$ results we found, and $k$ is usually about two to three times the number of the true causal anomalies \cite{tao2014metric, jarvelin2002cumulated}.
% nDCG則是選擇top-p個我們找到的結果，而p通常會略小於真正causal anomalies的數量一點點。
The nDCG is calculated using the top-$p$ results we found, and $p$ is usually slightly smaller than the number of the true causal anomalies.

%######################################################################

\subsection{Synthetic Data} \label{Synthetic Data}

\begin{figure}
    \centering
    \includegraphics[width=\columnwidth]{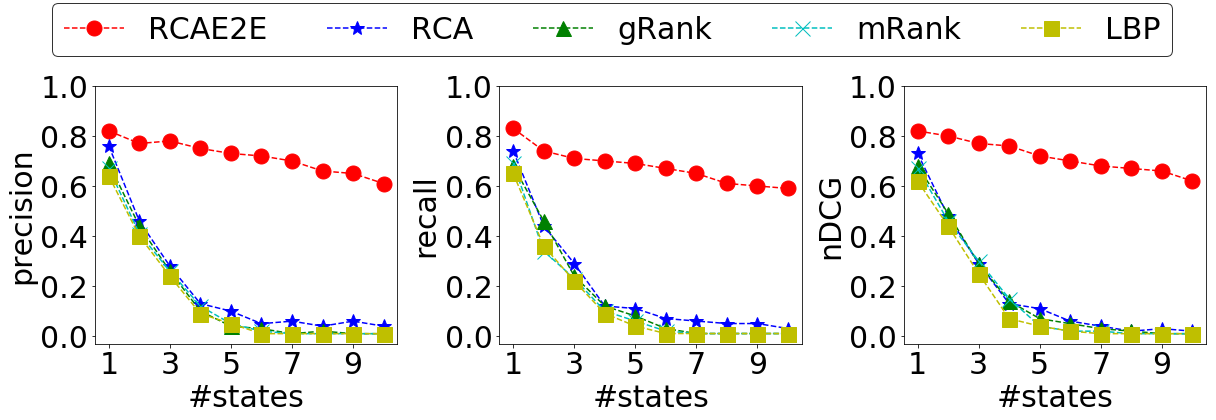}
    \caption{Performance vs. \#states}
    \label{01-states}
\end{figure}

\begin{figure}
    \centering
    \includegraphics[width=\columnwidth]{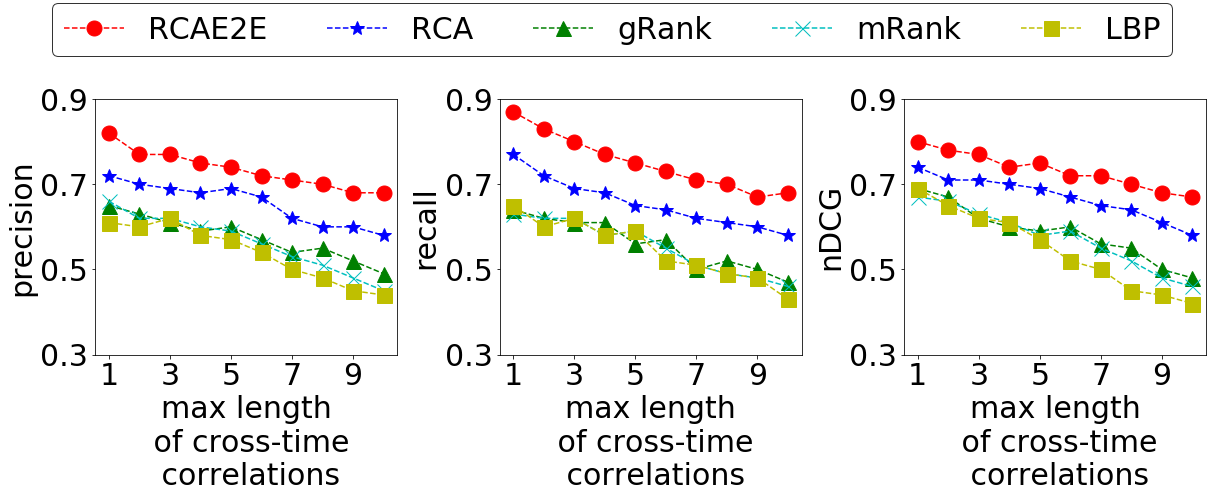}
    \caption{Performance vs. Max Length of Cross-Time Correlations}
    \label{02-max_length_of_cross_time_correlations}
\end{figure}

%######################################################################

\subsubsection{Data Generation} \label{Data Generation}

% 這邊的data generation分為兩個步驟：透過synthetic profile來產生run data，以及在run data裡面加上anomaly。
There are two steps in data generation: generate the run data based on the synthetic profile, and add anomalies into the run data.
% 首先我們會產生K個MRF，每個MRF都是透過Erdos-Renyi directed random graphs產生出來的。
First, we will generate $K$ MRFs, each of the MRF is generated based on Erd\H{o}s-R\'{e}nyi directed random graphs \cite{mohan2014node}.
% 再來我們會決定單一run data的cluster assignment。
Then we will determine the cluster assignment of single run data.
% 當我們有了MRFs以及cluster assignment，我們就有了所謂的synthetic profile。
When we have MRFs and cluster assignments, we have the so-called synthetic profile.
% 所以我們就可以透過這個synthetic profile來產生run data。
So we can generate the run data based on this synthetic profile.
% 至於在資料裡面插入anomalies，我們會先隨機選擇幾個個node使其為root cause，然後再利用Equation X算出相對應的b。
As for inserting anomalies into the data, we randomly select several nodes to make them the root cause anomalies, and then use Equation \ref{RCA_SCC Fault Propagation closed-form solution} to calculate the corresponding $b$.
% 我們利用amplitude-based anomaly generation的方式，透過b的值來調整data使其成為anomaly。
We use the amplitude-based anomaly generation method \cite{cheng2016ranking} to adjust the data to become anomalies by the value of $b$.

%---------------------------------------------------------------------------------------------------

% 這邊有兩個參數是動態的，其分別是cluster的數量(K)以及多個data point的window size(t_w)。
There are two parameters that are dynamic, which are the number of clusters ($K$) and the window size of multiple data points ($t\_w$).

%######################################################################

\subsubsection{Performance vs. \#states} \label{Performance vs. states}

% 在Figure X裡，我們可以很明顯地看到考慮state多元性的必要性。
In Figure \ref{01-states}, we can clearly see the necessity of considering the diversity of states.
% 這是因為如果他們只用單一一個模型來描述擁有多個state的end-to-end system，那麼他們所建立出來的correlation network會無法描述sensor在不同state的行為。
This is because if they use only a single model to describe an end-to-end system with multiple states, the correlation network they created would not be able to describe the behavior of sensors in different states.

\begin{figure}
    \centering
    \includegraphics[width=\columnwidth]{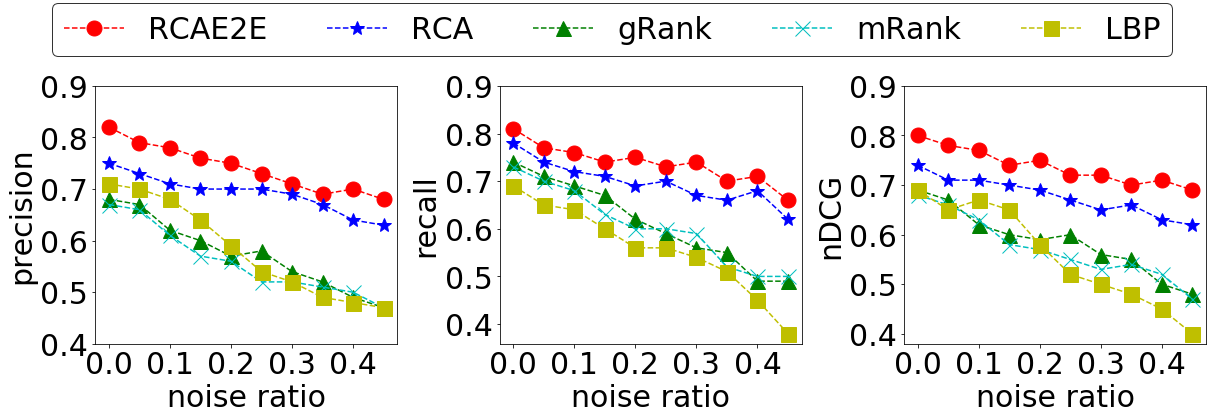}
    \caption{Performance vs. Noise Ratio}
    \label{03-noise_ratio}
\end{figure}

\begin{figure}
    \centering
    \includegraphics[width= 0.3\columnwidth]{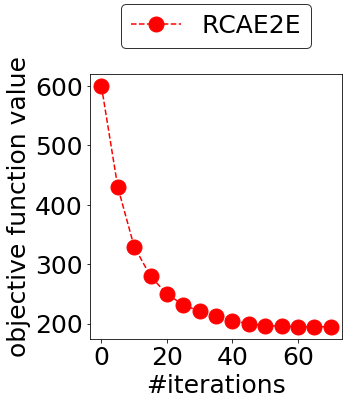}
    \caption{Convergence}
    \label{04-convergence}
\end{figure}

%######################################################################

\subsubsection{Performance vs. Max Length of Cross-Time Correlations} \label{Performance vs. Max Length of Cross-Time Correlations}

% 在Figure X裡，我們可以很明顯地看到把擁有不同time-lag的correlation分開考慮的必要性。
In Figure \ref{02-max_length_of_cross_time_correlations}, we can clearly see the necessity of separately considering the correlations with different time-lag.
% 當cross-time correlation的最大長度上升，所有方法的效能都會下降。
As the max length of cross-time correlations increases, the performance of all methods will drop.
% 但是因為RCAE2E有把擁有不同time-lag的correlation分開考慮，所以效能並不會像其他的方法一樣下降的那麼快。
But because RCAE2E separately considers the correlations with different time-lag, the performance does not drop as fast as other methods.

%######################################################################

\subsubsection{Noise Robustness} \label{Noise Robustness}

% 在Figure X裡，我們可以很明顯地看到考慮Fault Propagation的必要性。
In Figure \ref{03-noise_ratio}, we can clearly see the necessity of considering Fault Propagation.
% 當初Fault Propagation的假設就是認為不會透過correlations傳出去的anomalies就是noise。
The assumption of Fault Propagation is that the anomalies are noise if they won't propagate through the correlations.
% 也就是說，如果我們再資料裡面加入noise的話，有考慮Fault Propagation的方法(RCA以及RCAE2E)是不會受到太多影響的。
In other words, if we add noise to the data, the method of considering the Fault Propagation (RCA and RCAE2E) will not be affected too much.

%######################################################################

\subsubsection{Convergence} \label{Convergence}

% 在Figure X裡，我們可以很明顯地看到每次使用Equation X去更新都會使得objective function的值下降。
In Figure \ref{04-convergence}, we can clearly see that every time we use Equation (\ref{RCA_SCC Reconstruction Error updating rule}) to update the Problem (\ref{RCA_SCC Reconstruction Error objective function only s}), each update will monotonically decrease the value of the objective function.

%######################################################################

\subsection{Real-World Data} \label{Real-World Data}

%######################################################################

\subsubsection{Data Description} \label{Data Description}

% 這邊我們所用的真實世界資料是大型光電廠的機台的資料，而每個機台都包含了上百個感測器。
The real-world data we use here is the machine data of large photoelectric factory, and each machine contains hundreds of sensors.
% 我們有做一些基本的資料前處理，像是篩選資料，填補缺漏值，對資料做標準化。
We have done some basic data preprocessing, such as screening data, filling in missing values, and standardizing data.
% 除此之外，還有對資料做對齊(全部資料都應該要有一樣長的時間以及一樣的sensor名稱與數量)，以及對資料做downsample。
In addition, we have aligned the data (all data should have both the same length of timestamp and the same name and quantity of sensors), and down-sampled the data.
% 最終，我們餵到RCAE2E的data有以下的資料描述：run data的數量為136筆，單一個run data的時間長度為1460，sensor的數量為48個。
Finally, the data we fed to RCAE2E has the following data description: the number of run data is 136 ($R=136$), the length of single run data is 1460 ($T=1460$), and the number of sensors is 48 ($N=48$).

\begin{figure}
    \centering
    \includegraphics[width=0.9\columnwidth]{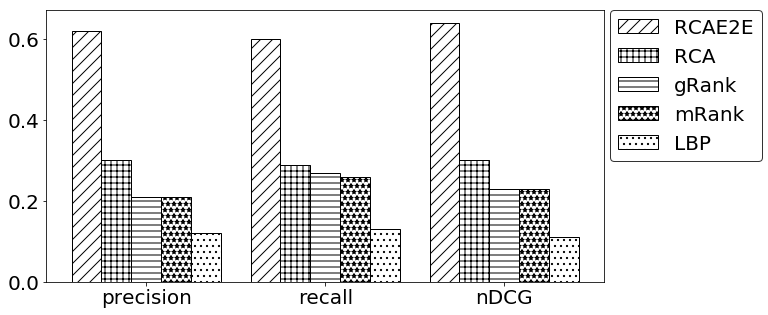}
    \caption{Performance}
    \label{01-performance}
\end{figure}

%######################################################################

\subsubsection{Performance} \label{Performance}

% 我們確認過有多個state存在在資料裡面，所以基本上使用的方法沒有考慮到state的多元性的話，其效能都會很差。
We have confirmed that there are multiple states in the data, so basically if the method does not take into account the diversity of the states, its performance will be very poor.
% 在Figure X裡，我們可以看到RCAE2E outperform其他方法，不管在precision、recall或者是nDCG。
In Figure \ref{01-performance}, we can see that RCAE2E outperform other methods whether in precision, recall, or nDCG.

%######################################################################

\subsubsection{RCA + Different Clustering Methods} \label{RCA + Different Clustering Methods}

\begin{figure}
    \centering
    \includegraphics[width=\columnwidth]{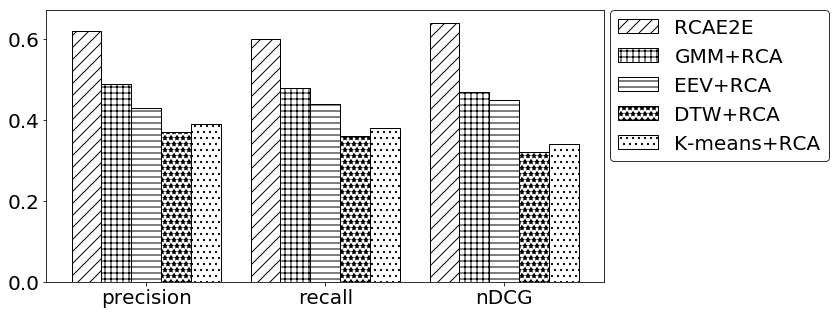}
    \caption{RCA + Different Clustering Methods}
    \label{02-RCA_different_clustering_methods}
\end{figure}

% RCA因為沒有考慮state的多元性，所以其性能在有多個state的data上面會非常差。
RCA does not consider the diversity of the states, so its performance will be very poor on data with multiple states.
% 我們讓RCA結合了一些常見的clustering的方法，可以看出在Figure X，其性能的確比原本的RCA還要好很多。
We combine RCA with some common clustering methods, and we can see that in Figure \ref{02-RCA_different_clustering_methods}, their performances are indeed much better than the original RCA.
% 但是在Figure X裡，我們可以看到RCAE2E的效能還是會贏過加上了clustering的RCA。
But in Figure \ref{02-RCA_different_clustering_methods}, we can see that RCAE2E still outperforms RCA with clustering.
\section{Conclusion} \label{Conclusion}

% 在這篇論文裡面，我們提出了一個稱為RCAE2E的framework，其是一個可以直接查找端到端系統中的異常原因的framework。
In this paper, we present a framework called \textit{Ranking Causal Anomalies in End-to-End System} (RCAE2E), which is a framework that can directly look for causal anomalies in the end-to-end system.
% 我們的貢獻是，我們提出來的RCAE2E徹底解決了RCA中沒有考慮多個states以及沒有把擁有不同time-lag的correlation分開考慮的問題。
Our contribution is that RCAE2E completely solve the problem of not considering the diversity of states and not separately considering the correlations with different time-lag in RCA.
% 總共有兩個主要的方法在RCAE2E裡，其分別是TICC_GTC以及RCA_SCC。
There are two main methods in RCAE2E, which are TICC\_GTC and RCA\_SCC.
% 最後，實驗的部分，我們驗證了假設的正確性以及存在性。
Finally, in the experimental part, we verify the correctness and existence of our method hypothesis.

\bibliographystyle{unsrt}
\bibliography{References/References.bib}

\end{document}